\documentclass[conference]{IEEEtran}
\IEEEoverridecommandlockouts
\usepackage[colorlinks=true, urlcolor=blue, linkcolor=red]{hyperref}
\usepackage{booktabs}
\usepackage{multirow} 
\usepackage{tabularx}
\usepackage{cite}
\usepackage{amsmath,amssymb,amsfonts}
\usepackage{algorithmic}
\usepackage{graphicx}
\usepackage{textcomp}
\usepackage{xcolor}
\usepackage{makecell}
\def\BibTeX{{\rm B\kern-.05em{\sc i\kern-.025em b}\kern-.08em
    T\kern-.1667em\lower.7ex\hbox{E}\kern-.125emX}}
\begin{document}

\title{PM-VIS\textsuperscript{+}: High-Performance Video Instance Segmentation without Video Annotation\\
}

\author{\IEEEauthorblockN{Zhangjing Yang, Dun Liu}
\IEEEauthorblockA{\textit{Nanjing Audit University} \\
Nanjing, China \\
yzj@nau.edu.cn, einstonliu@163.com}
\and
\and
\IEEEauthorblockN{Xin Wang}
\IEEEauthorblockA{\textit{University at Albany, SUNY} \\
NY, United States \\
xwang56@albany.edu}
\and
\IEEEauthorblockN{Zhe Li, Barathwaj Anandan, Yi Wu}
\IEEEauthorblockA{\textit{Lucid Motors} \\
Newark, CA, United States \\
\{lizoezhe, barathwajanandan, ywu.china\}@gmail.com}
}

\maketitle

\begin{abstract}
Video instance segmentation requires detecting, segmenting, and tracking objects in videos, typically relying on costly video annotations. This paper introduces a method that eliminates video annotations by utilizing image datasets. The PM-VIS algorithm is adapted to handle both bounding box and instance-level pixel annotations dynamically. We introduce ImageNet-bbox to supplement missing categories in video datasets and propose the PM-VIS\textsuperscript{+} algorithm to adjust supervision based on annotation types.
To enhance accuracy, we use pseudo masks and semi-supervised optimization techniques on unannotated video data. This method achieves high video instance segmentation performance without manual video annotations, offering a cost-effective solution and new perspectives for video instance segmentation applications. The code will be available in 
\href{https://github.com/ldknight/PM-VIS-plus} {https://github.com/ldknight/PM-VIS-plus}
\end{abstract}


\section{Introduction}
Video instance segmentation (VIS) aims to detect, segment, and track objects in videos. Since its introduction in 2019~\cite{VIS}, it has presented significant challenges and wide applications in video understanding, video editing, autonomous driving, and augmented reality. Advanced models rely on extensive video annotations for training~\cite{IDOL, GRAtt-VIS}. However, annotating video data, especially with object masks, is costly and time-consuming, making it difficult to scale existing methods~\cite{cheng2022pointly}. This paper reevaluates the need for semi-supervised VIS without relying on any video annotations.

Current instance segmentation models based on box annotations~\cite{discobox, boxinst} are designed for images and do not leverage temporal cues in videos, leading to lower accuracy when applied to video data. Videos contain rich contextual information and adhere to temporal consistency, where the same object across frames should have consistent labeling. This work leverages this constraint for mask learning in VIS.

Inspired by the PM-VIS algorithm~\cite{PMVIS}, we propose achieving high-precision VIS using only image datasets, addressing three key challenges:

\begin{itemize}
\item \textbf{Training on Image Datasets}: We adapt PM-VIS to train on image datasets by using single images as keyframes and generating reference frames through random cropping, thus simulating video training conditions.
\item \textbf{Category Matching}: We supplement VIS categories by introducing the ImageNet-bbox dataset, which includes all categories from existing video datasets or their parent categories.
\item \textbf{Dynamic Algorithm Adjustment}: We introduce the PM-VIS\textsuperscript{+} algorithm, which adjusts supervision loss based on the type of annotation, effectively utilizing both instance bounding box and pixel-level contour annotations.
\end{itemize}

Despite achieving VIS using image datasets, the model's recognition accuracy lags behind fully supervised methods due to dataset content differences, limited feature information in bounding box annotations, and the lack of temporal information richness.

To improve recognition rates, we propose a data optimization and model fine-tuning mechanism. First, we use the PM-VIS\textsuperscript{+} model to generate pseudo-labeled video data. Next, we optimize these pseudo-labels using the semi-supervised video object segmentation model DeAOT~\cite{Deaot} and a filtering mechanism. Finally, we train on the optimized video data with the PM-VIS\textsuperscript{+} algorithm.

The contributions of this work are as follows:
\begin{itemize}
\item Proposing a method to achieve VIS using only image dataset annotations, leveraging ImageNet-bbox and COCO datasets.
\item Introducing a pseudo-label filtering mechanism to remove noise from pseudo-labels.
\item Developing the PM-VIS\textsuperscript{+} algorithm, which dynamically adjusts supervision based on the type of data annotation, using weakly supervised and fully supervised losses for instance bounding boxes and pixel-level contours.
\end{itemize}

\section{Related work}
\subsection{Video instance segmentation}
VIS methods are divided into offline and online types. Offline methods~\cite{heo2022vita,wu2022seqformer} analyze videos with future frames during inference, suitable for tasks like video editing. Early methods~\cite{lin2021video,bertasius2020classifying} used mask propagation for tracking. With DETR~\cite{carion2020end}, query-based VIS methods~\cite{heo2022vita,wu2022seqformer} became popular.
Online methods~\cite{IDOL,VIS} handle segmentation and tracking by embedding similarity and optimizing results, used in surveillance and autonomous driving. Mask-Track R-CNN~\cite{VIS} extends Mask R-CNN for tracking. IDOL~\cite{IDOL} uses contrastive learning on instance queries, surpassing offline models. MinVIS~\cite{huang2022minvis} and GenVIS~\cite{heo2023generalized} improve performance with Mask2Former~\cite{cheng2022masked}. Our semi-supervised VIS algorithm will use IDOL as the baseline.


MaskFreeVIS~\cite{ke2023mask} uses bounding box supervision for VIS, incorporating a model~\cite{boxinst} and achieving competitive performance with temporal or spatial losses. FlowIRN~\cite{liu2021weakly} uses classification labels and optical flow but has limited performance. Previous weakly supervised methods~\cite{ke2023mask,liu2021weakly} face challenges due to insufficient data optimization.

\begin{figure*}[!h]
	\centering
	\scalebox{0.2}{\includegraphics{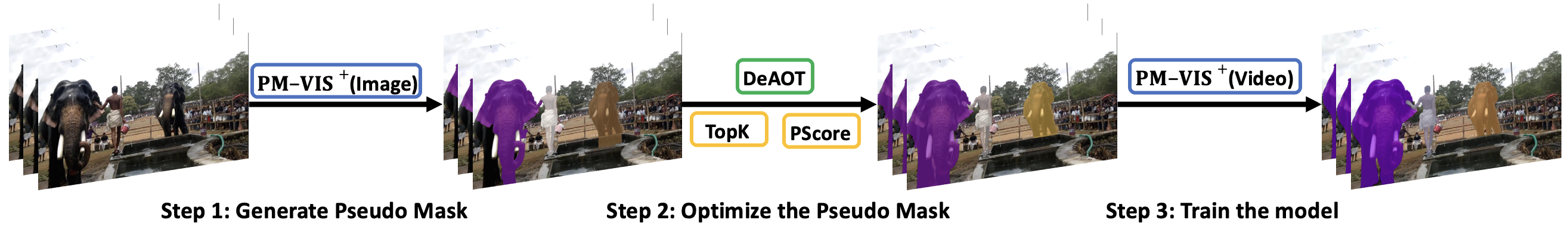}}
	\caption{Method flow diagram.}
	\label{fig:methods_process} 
\end{figure*}
\section{methodology}
\subsection{Method flow}
Fig.~\ref{fig:methods_process} illustrates the workflow of PM-VIS\textsuperscript{+} without using any video annotation data, where PM-VIS\textsuperscript{+}(Image) and PM-VIS\textsuperscript{+}(Video) respectively denote the PM-VIS\textsuperscript{+} models trained on image data and pseudo-labeled video data.

The method consists of three steps: generating pseudo-labeled data, optimizing pseudo-labeled data, and training the model using pseudo-labeled data. Specifically:
\begin{enumerate}
\item Generating Pseudo-labeled Data: The VIS model PM-VIS\textsuperscript{+}(Image), trained on image datasets, infers the video data to obtain pseudo-labeled video data. Although each video data obtains pseudo-labels after model prediction, the quality of these pseudo-labels varies. Excessive misleading data can significantly hinder model learning, and even if reused for subsequent model training, recognition accuracy may not improve.

\item Optimizing Pseudo-labeled Data: The obtained pseudo-labeled data undergoes optimization using the semi-supervised video object segmentation algorithm DeAOT [34], along with filtering based on prediction scores. While optimization through tracking methods improves the overall quality of the data by addressing issues such as missing, erroneous, or inaccurate instance pseudo-masks, it still cannot be directly used as training data for the next stage. The main reason is that while the overall quality of the data has improved, the true class of instances cannot be determined. Therefore, this paper filters the data based on the number of instances in the video and the average prediction score to mitigate the impact of erroneous data to a certain extent.

\item Training the Model Using Pseudo-labeled Data: The optimized pseudo-labeled data is used to train the PM-VIS\textsuperscript{+}(Video) model. After the first two steps, data with fewer errors and higher overall quality is obtained. Hence, this paper proposes training the PM-VIS\textsuperscript{+} model on pseudo-labeled video data to learn potential information within the video data.
\end{enumerate}

Through these three steps, this paper achieves a certain recognition rate of VIS models without using any manually annotated video data. Compared to conventional methods that directly use manually annotated video datasets for algorithm training, this paper introduces pseudo-labeled data generation and optimization steps, along with the use of the ImageNet-bbox dataset containing instance bounding box annotations.

\subsection{Model training process}
\begin{figure}[!h]
	\centering
	\scalebox{0.2}{\includegraphics{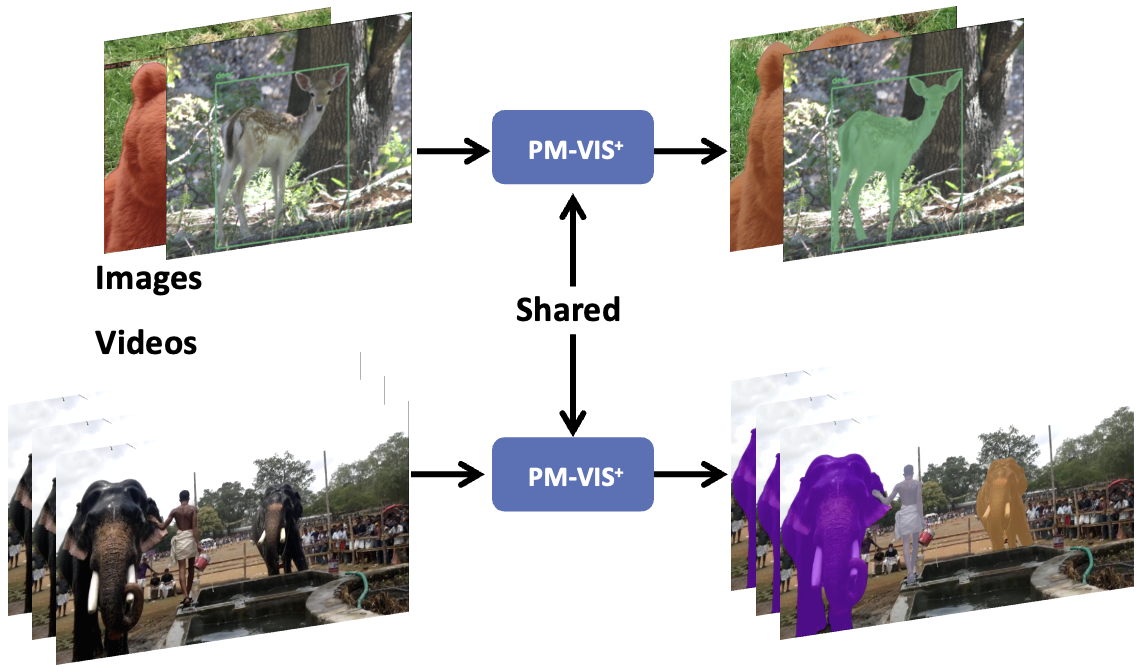}}
	\caption{Model training process.}
	\label{fig:training_process} 
\end{figure}

\hfill 

Fig.~\ref{fig:training_process} illustrates the training process of the proposed PM-VIS\textsuperscript{+} model, which involves training the model on two types of data under different supervision conditions. In this process, image data is used as auxiliary data to train the VIS model corresponding to the target dataset categories. Notably, in addition to using COCO data with pixel-level annotations, the model also utilizes ImageNet-bbox data, which only contains bounding box annotations. To address the categories not covered by COCO but present in video data such as the 40 categories in YTVIS2019\cite{VIS}, the model incorporates additional data from ImageNet-bbox.

The proposed PM-VIS\textsuperscript{+} model dynamically adjusts the supervision signal based on the type of annotation information available in the training data. Specifically, leveraging the characteristics of IDOL, the COCO dataset with instance-level pixel annotations is directly used for training. In contrast, the ImageNet-bbox dataset, which lacks pixel-level annotations and only provides bounding boxes, cannot supervise the mask prediction head during training. Therefore, the model trains the mask prediction head only on the COCO dataset while retaining instance-level bounding box supervision on the ImageNet-bbox dataset, as shown in the first row of Fig.~\ref{fig:training_process}.

It is important to note that although the model trained on the image datasets can perform VIS, its recognition accuracy is relatively weak. Therefore, the model is used to generate pseudo-labels for the video data, which are then optimized to further enhance the model. For video data with pseudo-labels, despite having pixel-level instance annotations, the quality of these annotations varies. Relying solely on pixel-level supervision is insufficient to improve the algorithm significantly. Hence, the PM-VIS\textsuperscript{+} algorithm employed for training the video data with pseudo-labels retains both the BoxInstLoss\cite{boxinst} for bounding box supervision and the MaskLoss\cite{DiceLoss,FocalLoss} for pixel-level instance segmentation supervision.

\subsection{Video pseudo-label data optimization strategy}

\begin{figure}[!h]
	\centering
	\scalebox{0.62}{\includegraphics{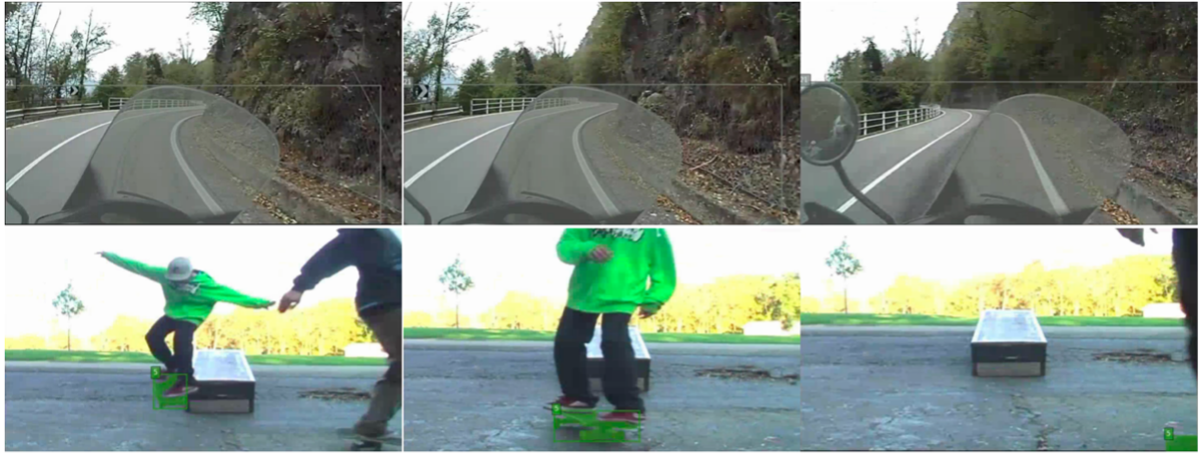}}
	\caption{PM-VIS\textsuperscript{+}(Image) visualization of missed detection data relative to real data.}
	\label{fig:missingdata} 
\end{figure}

\begin{figure}[!h]
	\centering
	\scalebox{0.105}{\includegraphics{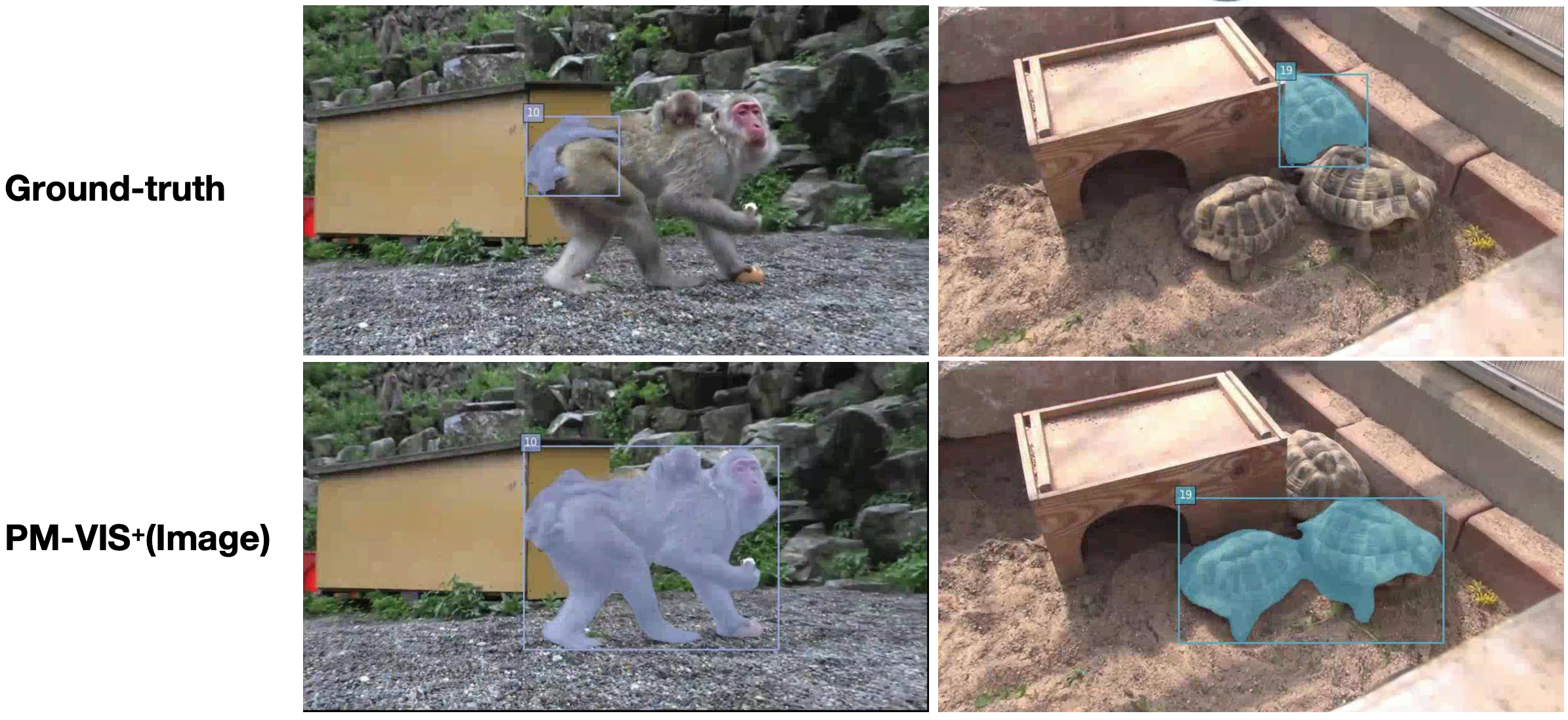}}
	\caption{PM-VIS\textsuperscript{+}(Image) visualization of reasoning results.}
	\label{fig:visibleRes} 
\end{figure}

As shown in Fig.~\ref{fig:missingdata} and Fig.~\ref{fig:visibleRes}, the PM-VIS\textsuperscript{+} VIS model trained only on image data demonstrates poor recognition capabilities, resulting in numerous errors and missed detections compared to using actual annotated data. The visualization issues can be categorized into three types: missed instances, false positives, and inaccurate instance detections. This study attributes these problems primarily to the different data sources. However, retraining the PM-VIS\textsuperscript{+} model with diverse, accurately segmented, and correctly categorized data can significantly improve the accuracy of VIS.

To achieve this, the study employs the semi-supervised video object segmentation model DeAOT for data optimization. The optimization process consists of two steps:
\begin{enumerate}
\item Initialization: The pseudo-labels predicted by PM-VIS\textsuperscript{+} (Image) are used to initialize the DeAOT model, which then tracks the instance masks throughout the video. Unlike conventional semi-supervised video object segmentation, which initializes using the first frame's instance mask, this study determines the initialization frame based on the highest prediction score of the instance in the current video. This initialization frame is referred to as the keyframe.

\item Tracking: Starting from the keyframe, DeAOT tracks the instances to both ends of the video (i.e., towards the first and last frames), and the results are merged to obtain the complete pseudo-mask for the instance across the video.
\end{enumerate}

This tracking optimization addresses issues of missed instances and inaccurate instance segmentation. However, it does not fully resolve the problem of false positives. Therefore, the optimized data cannot be directly used for model training.

To address this, the study proposes two pseudo-label data filtering strategies: TopK and PScore. Specifically:

\begin{enumerate}
\item TopK Filtering: Based on the average score of the instance in the video predicted by the PM-VIS\textsuperscript{+} (Image) model, the TopK method selects the top K highest-scoring instance pseudo-masks, ignoring the rest.

\item PScore Filtering: Using the average score of the instance in the video predicted by the PM-VIS\textsuperscript{+} (Image) model, the PScore method applies a score threshold $\tau$ to filter the pseudo-label data. Only instances with an average score above the threshold are retained, while the rest are discarded.
\end{enumerate}

Through these data optimization strategies, the improved pseudo-label data shows enhancements in instance segmentation quantity, segmentation quality, and instance category prediction accuracy.

\section{Experiment}
\subsection{Datasets}
The proposed semi-supervised VIS method, PM-VIS\textsuperscript{+}, requires not only the COCO dataset, which contains instance-level pixel-wise contour annotations, but also the ImageNet-bbox dataset\cite{su2012crowdsourcing}, which provides bounding box annotations for objects, as auxiliary data. In order to further improve the algorithm's recognition rate, this chapter also utilizes the video datasets YTVIS2019\cite{VIS}, YTVIS2021\cite{YTVIS2021}, and OVIS\cite{OVIS}, without the need for manual annotation.

\begin{table}[htbp]
\caption{Statistics on the category proportions of image data sets and video data sets}
\begin{center}
\begin{tabular}{|c|c|c|c|c|}
\hline
\multicolumn{2}{|c|}{\textbf{Name}} & \textbf{\makecell{YTVIS2019\\(40)}} & \textbf{\makecell{YTVIS2021\\(40)}} & \textbf{\makecell{OVIS\\(25)}} \\ 
\hline
\multirow{3}{*}{COCO} & \textbf{\textit{Class}} & 21 / 19 & 23 / 21 & 17 / 15 \\
& \textbf{\textit{Images}} & 86646 & 88462 & 86116 \\
& \textbf{\textit{Anno}} & 420670 & 434158 & 405865 \\ 
\hline
\multirow{3}{*}{\makecell{ImageNet\\-bbox}} & \textbf{\textit{Class}} & 21 / 21 & 19 / 19 & 12 / 10 \\
& \textbf{\textit{Images}} & 7178 & 6768 & 5814 \\
& \textbf{\textit{Anno}} & 8101 & 7651 & 6783 \\ 
\hline
\end{tabular}
\label{table:categoryproportions}
\end{center}
\end{table}

Table~\ref{table:categoryproportions} shows the distribution of categories in both image and video datasets. The numbers in parentheses after each dataset indicate the current number of categories in that dataset. For instance, ``YTVIS2019(40)" indicates that the YTVIS2019 dataset comprises 40 categories. It can be observed that among the video datasets YTVIS2019, YTVIS2021, and OVIS, there are 21, 19, and 15 categories respectively that overlap with the COCO dataset. The remaining categories are supplemented from ImageNet-bbox. However, it was found in practical operation that for video datasets, there may exist cases where one category corresponds to multiple categories in COCO. For example, the entry ``21 / 19" in the table indicates that 21 categories from the COCO dataset match with the 19 categories in YTVIS19. For instance, the ``snowboard" category in YTVIS19 corresponds to both ``skis" and "snowboard" categories in COCO.

In addition to the COCO dataset, this paper also utilizes the ImageNet-bbox dataset, which contains annotations for 3000 categories, including bounding boxes for objects, to supplement categories that cannot be filled in with COCO. As shown in Table~\ref{table:categoryproportions}, the 21 categories from YTVIS2019, 19 categories from YTVIS2021, and 21 categories from OVIS will be supplemented using ImageNet-bbox.

From the table, it is evident that in terms of data volume, the quantity from ImageNet-bbox is noticeably less than that from COCO. Additionally, in terms of data standardization, relative to the COCO instance segmentation dataset, the standardization of ImageNet-bbox data is poorer, primarily reflected in the mismatch between annotated data quantity and the actual downloaded image data. Taking the example of the image data corresponding to the YTVIS2019 dataset, there are 41 images that are not found in the downloaded image set and need to be supplemented by searching from the ImageNet-22K image set. Furthermore, there are 11 images for which corresponding image sources cannot be found, leading to their exclusion from the experiment, resulting in a final set of 7178 images across 21 categories. Consequently, there may be issues in experimental results where the category data trained from ImageNet-bbox do not predict accurately, leading to lower model recognition rates.

\subsection{Experimental setup}
For VIS algorithms, unless otherwise specified, this paper employs the same hyperparameters as IDOL and continues to use Detectron2\cite{wu2019detectron2} for related experiments. All VIS models in this paper are pretrained on the COCO dataset. 

\subsection{ablation experiment}
This chapter will conduct ablation experiments on the proposed algorithm to validate the roles of the components proposed in this paper. The ablation experiments here will be evaluated on the validation set of YTVIS2019, using ResNet-50 as the backbone network.

\subsubsection{Training different experimental configurations of the PM-VIS\textsuperscript{+} (Image) model on the image dataset}

\begin{table}[htbp]
    \caption{Experimental plan for training PM-VIS\textsuperscript{+} model on image data}
    \begin{center}
    \renewcommand\arraystretch{1.3}
    \setlength{\tabcolsep}{3pt}
    \begin{tabular}{|c|c|c|c|c|c|c|}
        \hline
        \textbf{MaskLoss} & \textbf{BoxInstLoss} & \textbf{AP} & \textbf{AP50} & \textbf{AP75} & \textbf{AR1} & \textbf{AR10} \\
        \hline
        $\checkmark$ & $\times$ & \textbf{41.9} & \textbf{64.9} & \textbf{45.6} & 40.1 & 50.0 \\
        $\checkmark$ & $\checkmark$ & 40.4 & 61.7 & 43.2 & \textbf{40.9} & \textbf{50.1} \\
        $\checkmark$ & ImageNet-bbox & 40.3 & 63.1 & 41.4 & 40.2 & 49.6 \\
        Freezing Mask Head & $\times$ & 40.5 & 62.2 & 43.2 & 39.8 & 50.2 \\
        $\times$ & $\checkmark$ & 37.3 & 59.7 & 39.6 & 36.6 & 45.3 \\
        \hline
    \end{tabular}
    \label{table:PM-VISImage}
    \end{center}
\end{table}


Table~\ref{table:PM-VISImage} presents the performance of the proposed PM-VIS\textsuperscript{+} (Image) algorithm trained with different experimental configurations on the image dataset. It can be observed that the BoxInstLoss does not effectively supervise the model on the image dataset. When using this loss for supervision, the model accuracy is 40.4\%, while without using this loss, the accuracy increases by 1.5\% to reach 41.9\%.

Furthermore, this paper also compares the impact of BoxInstLoss on two different image datasets. It is found that solely using BoxInstLoss on the ImageNet-bbox dataset results in a model accuracy of 40.3\%, which is a decrease of 0.1\% compared to using it on both image datasets, which achieved 40.4\%. Thus, BoxInstLoss may have a slight promoting effect on the training of the COCO dataset but is relatively minimal.

It is evident that even when the mask prediction head of the model is frozen during training, the model still possesses certain instance segmentation capabilities, with an accuracy of 40.5\%. This is mainly attributed to the pretrained model trained on the COCO dataset. When the instance mask prediction head of the model is unfrozen, the model's average precision (AP) increases by 1.4\% to reach 41.9\%. Therefore, the ImageNet-bbox image dataset not only supplements the categories but also enhances the model's recognition capabilities.

However, if only the ImageNet-bbox image dataset is used for training, the model's recognition capability decreases to 37.3\%. This indicates the crucial importance of pixel-level annotation information present in COCO, while the facilitating effect of ImageNet-bbox is limited.

\subsubsection{The impact of the hyperparameter K of the TopK filtering method for pseudo-labeled video data on the recognition rate of PM-VIS\textsuperscript{+} (Video)}


\begin{table}[htbp]
\caption{The impact of the hyperparameter K of the video pseudo-label data filtering method TopK on the model recognition rate}
\begin{center}
\begin{tabular}{|c|c|c|c|c|c|}
\hline
\textbf{K} & \textbf{AP} & \textbf{AP50} & \textbf{AP75} & \textbf{AR1} & \textbf{AR10} \\ 
\hline
0 & 41.2 & 61.3 & 44.4 & 41.9 & 53.3 \\
1 & 36.9 & 55.1 & 39.4 & 37.8 & 46.7 \\
2 & \textbf{44.7} & 67.4 & 46.5 & 43.0 & 54.7 \\
3 & 44.6 & 67.1 & \textbf{48.3} & \textbf{43.9} & 55.2 \\
4 & \textbf{44.7} & \textbf{67.5} & 47.4 & 43.8 & \textbf{56.1} \\
5 & 43.1 & 65.1 & 46.9 & \textbf{43.9} & 56.0 \\
\hline
\end{tabular}
\label{table:TopK}
\end{center}
\end{table}

Table~\ref{table:TopK} illustrates the impact of the hyperparameter K of the TopK filtering method on the recognition rate of PM-VIS\textsuperscript{+}(Video). Here, the hyperparameter $\tau$ of PScore is set to 0.2 by default. It is worth noting that these pseudo-labeled data are derived from the inference results of the PM-VIS\textsuperscript{+} (Image) model trained on the ImageNet-bbox and COCO datasets on the YTVIS2019 training set, followed by refinement using the DeAOT model.

It can be observed that the model's recognition capability, when not using this filtering mechanism, decreases by 0.7\% relative to PM-VIS\textsuperscript{+}(Image), reaching 41.2\%. This indicates the presence of considerable noisy data within the video data containing pseudo-labels, which, if directly used for training, can mislead the model. When the TopK strategy with K set to 4, the model can effectively learn information from the pseudo-labeled video data, resulting in a relative improvement of 3.5\% in model AP compared to not using this filtering method, achieving the optimal performance of 44.7\%.

\subsubsection{The impact of the hyperparameter $\tau$ of the PScore filtering method on the recognition rate of PM-VIS\textsuperscript{+} (Video)}


\begin{table}[htbp]
\caption{The impact of different supervision signals on the pseudo-label model PM-VIS\textsuperscript{+}(Video)}
\begin{center}
\begin{tabular}{|c|c|c|c|c|c|c|}
\hline
\textbf{MaskLoss} & \textbf{BoxInstLoss} & \textbf{AP} & \textbf{AP50} & \textbf{AP75} & \textbf{AR1} & \textbf{AR10} \\ 
\hline
\checkmark & \checkmark & \textbf{44.7} & \textbf{67.4} & \textbf{46.5} & \textbf{43.0} & \textbf{54.7} \\
\checkmark & $\times$ & 43.3 & 66.8 & 44.8 & 43.2 & 54.8 \\
$\times$ & \checkmark & 40.2 & 65.4 & 42.4 & 40.6 & 52.2 \\
$\times$ & $\times$ & 38.4 & 62.6 & 39.1 & 39.8 & 49.4 \\
\hline
\end{tabular}
\label{table:supervisionsignals}
\end{center}
\end{table}

Table~\ref{table:PScore} presents the impact of the PScore filtering method with different $\tau$ values on the recognition rate of the PM-VIS\textsuperscript{+} (Video) model trained on filtered pseudo-labeled video data. Here, the K value of the TopK filtering method is set to 4 by default. Similar to the TopK filtering method, the pseudo-labeled data used here are derived from the inference results of the PM-VIS\textsuperscript{+} (Image) model on videos, followed by refinement using the DeAOT model.

It can be observed that when using the PScore filtering method, the model accuracy increases by 1.9\% relative to not using this method, reaching 42.8\%. It is evident that there is a significant amount of erroneous information in the predicted pseudo-labels, and solely using the TopK filtering method cannot effectively filter out the noise data. For the PScore filtering method, setting $\tau$ to 0.2, the model accuracy reaches 44.7\% when trained on the filtered pseudo-labeled data. Therefore, using PScore effectively removes noise data from the pseudo-labeled data, improves the overall quality of the data, and enhances the model's recognition capability.

\subsubsection{The impact of different supervision signals on the pseudo-labeled model PM-VIS\textsuperscript{+} (Video)}

\begin{table}[htbp]
\caption{The impact of different supervision signals on the pseudo-label model PM-VIS\textsuperscript{+}(Video)}
\begin{center}
\begin{tabular}{|c|c|c|c|c|c|c|}
\hline
\textbf{MaskLoss} & \textbf{BoxInstLoss} & \textbf{AP} & \textbf{AP50} & \textbf{AP75} & \textbf{AR1} & \textbf{AR10} \\ 
\hline
\checkmark & \checkmark & \textbf{44.7} & \textbf{67.4} & \textbf{46.5} & \textbf{43.0} & \textbf{54.7} \\
\checkmark & $\times$ & 43.3 & 66.8 & 44.8 & 43.2 & 54.8 \\
$\times$ & \checkmark & 40.2 & 65.4 & 42.4 & 40.6 & 52.2 \\
$\times$ & $\times$ & 38.4 & 62.6 & 39.1 & 39.8 & 49.4 \\
\hline
\end{tabular}
\label{table:supervisionsignals}
\end{center}
\end{table}

Table~\ref{table:supervisionsignals} displays the results of training different architectures of the PM-VIS\textsuperscript{+} (Video) algorithm on pseudo-labeled video data. The pseudo-labeled data used here have been filtered using both the TopK and PScore methods.
It can be observed that both MaskLoss and BoxInstLoss contribute to the model's learning of information from pseudo-labeled data. When using only MaskLoss or BoxInstLoss, the model's recognition capability decreases by 1.4\% and 4.5\%, respectively, compared to when both are used. The magnitude of the decrease indicates that the effect of MaskLoss is significantly stronger than the benefits brought by BoxInstLoss. When both MaskLoss and BoxInstLoss are used as supervision signals simultaneously, the model's recognition capability reaches its optimal performance, achieving an AP of 44.7\%.

When using only MaskLoss without supervision from BoxInstLoss, there is a significant decrease in the algorithm's recognition capability. This is mainly because the confidence of the instance pseudo-masks in the pseudo-labeled data is relatively low, indicating varying qualities of instance pixel-level contours. Simply trusting the pseudo-labeled pixel-level segmentation results cannot enhance the algorithm's performance.


\subsubsection{The impact of the quality of data in earlier stages on the final accuracy of the VIS model}

\begin{table}[htbp]
\caption{The impact of PM-VIS\textsuperscript{+}(Image) model recognition ability on the accuracy of the final VIS model}
\begin{center}
\begin{tabular}{|p{13.165em}|c|p{7.25em}|c|}
\hline
\textbf{Method} & \textbf{AP} & \textbf{Method} & \textbf{AP} \\
\hline
PM-VIS\textsuperscript{+}(Image)+BoxInstLoss & 40.4 & \multirow{2}{*}{PM-VIS\textsuperscript{+}(Video)} & 44.0 \\
PM-VIS\textsuperscript{+}(Image) & \textbf{41.9} & & \textbf{44.7} \\
\hline
\end{tabular}
\label{table:impactfinalvis}
\end{center}
\end{table}

Table~\ref{table:impactfinalvis} illustrates the impact of pseudo-masks generated in the previous stage on the recognition capability of the final VIS model. It is noticeable that there is a 1.5\% difference in accuracy between the two PM-VIS\textsuperscript{+}(Image) models on the validation set. Based on this difference, even until the final model trained using pseudo-labeled video masks, there remains a 0.7\% AP gap in recognition capability. Therefore, it can be concluded that the recognition capability of the PM-VIS\textsuperscript{+}(Image) model will significantly affect the quality of subsequent video data pseudo-labels, consequently influencing the recognition rate of the PM-VIS\textsuperscript{+}(Video) model.





Therefore, the quality of data at each stage, from auxiliary image data to the quality of pseudo-labels, will impact the method's final recognition accuracy. Optimal selection of initialization image data, designing appropriate optimization and filtering mechanisms, and utilizing superior PM-VIS\textsuperscript{+} (Image) models will be crucial for experimental results.

\subsection{Comparison of data set effects}
In this section, we conduct experiments on the YTVIS2019, YTVIS2021, and OVIS datasets using both ResNet-50\cite{ResNet-50} and Swin-L\cite{Swin-L} backbone networks for the proposed method PM-VIS\textsuperscript{+}, comparing its recognition accuracy with other methods.

\subsubsection{ResNet-50 backbone network}

\begin{table}[htbp]
\small
\centering
\renewcommand\arraystretch{1.3}
\setlength\tabcolsep{2.2pt} 
\caption{Comparison of algorithm effects using ResNet-50 backbone network on YTVIS2019}
\begin{tabular}{|c|c|c|c|c|c|c|c|}
\hline
\textbf{Method} & \textbf{Type} & \textbf{Sup.} & \textbf{AP} & \textbf{AP50} & \textbf{AP75} & \textbf{AR1} & \textbf{AR10} \\
\hline
MaskFreeVIS & Video & Box & 46.6 & 72.5 & 49.7 & 44.9 & 55.7 \\
IDOL-BoxInst & Video & Box & 43.9 & 71.0 & 47.8 & 42.9 & 52.7 \\
PM-VIS & Video & Box & 48.7 & 73.4 & 52.4 & 45.2 & 55.3 \\
PM-VIS\textsuperscript{+} & Image & Pixel/Box & 41.9 & 64.9 & 45.6 & 40.1 & 50.0 \\
PM-VIS\textsuperscript{+} & Video & - & \textbf{44.7} & \textbf{67.5} & \textbf{47.4} & \textbf{43.8} & \textbf{56.1} \\
\hline
\end{tabular}
\label{tab:ResNet-50YTVIS2019}
\end{table}


\begin{table}[htbp]
\small
\centering
\renewcommand\arraystretch{1.3}
\setlength\tabcolsep{2.2pt} 
\caption{Comparison of algorithm effects using ResNet-50 backbone network on YTVIS2021}
\begin{tabular}{|c|c|c|c|c|c|c|c|}
\hline
\textbf{Method} & \textbf{Type} & \textbf{Sup.} & \textbf{AP} & \textbf{AP50} & \textbf{AP75} & \textbf{AR1} & \textbf{AR10} \\
\hline
MaskFreeVIS & Video & Box & 40.9 & 65.8 & 43.3 & 37.1 & 50.5 \\
IDOL-BoxInst & Video & Box & 41.8 & 67.4 & 43.5 & 36.5 & 50.3 \\
PM-VIS & Video & Box & 44.6 & 69.5 & 49.0 & 38.9 & 52.1 \\
PM-VIS\textsuperscript{+} & Image & Pixel/Box & 37.6 & 59.9 & 39.1 & 35.6 & 47.7 \\
PM-VIS\textsuperscript{+} & Video & - & \textbf{39.7} & \textbf{61.1} & \textbf{41.6} & \textbf{37.3} & \textbf{52.8} \\
\hline
\end{tabular}
\label{tab:ResNet-50YTVIS2021}
\end{table}


\begin{table}[htbp]
\small
\centering
\renewcommand\arraystretch{1.3}
\setlength\tabcolsep{2.2pt} 
\caption{Comparison of algorithm effects using ResNet-50 backbone network on OVIS}
\begin{tabular}{|c|c|c|c|c|c|c|c|}
\hline
\textbf{Method} & \textbf{Type} & \textbf{Sup.} & \textbf{AP} & \textbf{AP50} & \textbf{AP75} & \textbf{AR1} & \textbf{AR10} \\
\hline
MaskFreeVIS & Video & Box & 15.7 & 35.1 & 13.1 & 10.1 & 20.4 \\
IDOL-BoxInst & Video & Box & 25.4 & 47.4 & 23.7 & 12.9 & 32.7 \\
PM-VIS & Video & Box & 27.8 & 48.5 & 27.4 & 13.6 & 36.0 \\
PM-VIS\textsuperscript{+} & Image & Pixel/Box & 12.5 & 25.8 & 11.2 & 8.8 & 20.2 \\
PM-VIS\textsuperscript{+} & Video & - & \textbf{15.1} & \textbf{31.3} & \textbf{13.4} & \textbf{10.4} & \textbf{22.5} \\
\hline
\end{tabular}
\label{tab:ResNet-50OVIS}
\end{table}

Table~\ref{tab:ResNet-50YTVIS2019} ,Table~\ref{tab:ResNet-50YTVIS2021} and Table~\ref{tab:ResNet-50OVIS} respectively present the performance of the proposed semi-supervised VIS method PM-VIS\textsuperscript{+} based on image data on the YTVIS2019, YTVIS2021, and OVIS datasets. It can be observed that on all three datasets, the recognition accuracy of the PM-VIS\textsuperscript{+} (Image) model trained solely on image data achieves satisfactory results.
However, when using PM-VIS\textsuperscript{+} (Video) trained on the optimized YTVIS2019, YTVIS2021, and OVIS datasets, the model's recognition capability relative to PM-VIS\textsuperscript{+} (Image) improves by 2.8\%, 2.1\%, and 2.6\%, reaching 44.7\%, 39.7\%, and 15.1\% respectively. Therefore, it can be concluded that the proposed method not only achieves a certain level of recognition effectiveness on weaker backbone networks but also significantly improves through optimization. In certain scenarios, the effectiveness of the proposed method exceeds that of methods based on instance bounding boxes.

\subsubsection{Swin-L backbone network}

\begin{table}[htbp]
\small
\centering
\renewcommand\arraystretch{1.3}
\setlength\tabcolsep{2.2pt} 
\caption{Comparison of algorithm effects using Swin-L backbone network on YTVIS2019}
\begin{tabular}{|c|c|c|c|c|c|c|c|}
\hline
\textbf{Method} & \textbf{Type} & \textbf{Sup.} & \textbf{AP} & \textbf{AP50} & \textbf{AP75} & \textbf{AR1} & \textbf{AR10} \\
\hline
MaskFreeVIS & Video & Box & 55.3 & 82.5 & 60.8 & 55.3 & 82.5 \\
IDOL-BoxInst & Video & Box & 56.5 & 83.3 & 64.1 & 56.5 & 83.3 \\
PM-VIS & Video & Box & 59.7 & 84.8 & 67.7 & 59.7 & 84.8 \\
PM-VIS\textsuperscript{+} & Image & Pixel/Box & 54.5 & 77.7 & 60.6 & 48.9 & 60.6 \\
PM-VIS\textsuperscript{+} & Video & - & \textbf{57.2} & \textbf{78.7} & \textbf{63.9} & \textbf{50.4} & \textbf{63.4} \\
\hline
\end{tabular}
\label{tab:swinlytvis2019}
\end{table}


\begin{table}[htbp]
\small
\centering
\renewcommand\arraystretch{1.3}
\setlength\tabcolsep{2.2pt} 
\caption{Comparison of algorithm effects using Swin-L backbone network on YTVIS2021}
\begin{tabular}{|c|c|c|c|c|c|c|c|}
\hline
\textbf{Method} & \textbf{Type} & \textbf{Sup.} & \textbf{AP} & \textbf{AP50} & \textbf{AP75} & \textbf{AR1} & \textbf{AR10} \\
\hline
IDOL-BoxInst & Video & Box & 53.2 & 79.7 & 59.6 & 43.0 & 58.1 \\
PM-VIS & Video & Box & 55.8 & 80.6 & 61.8 & 44.4 & 60.2 \\
PM-VIS\textsuperscript{+} & Image & Pixel/Box & 50.0 & 72.2 & 55.5 & 42.5 & 56.3 \\
PM-VIS\textsuperscript{+} & Video & - & \textbf{52.9} & \textbf{78.5} & \textbf{56.6} & \textbf{43.7} & \textbf{59.9} \\
\hline
\end{tabular}
\label{tab:swinlytvis2021}
\end{table}


\begin{table}[htbp]
\small
\centering
\renewcommand\arraystretch{1.3}
\setlength\tabcolsep{2.2pt} 
\caption{Comparison of algorithm effects using Swin-L backbone network on OVIS}
\begin{tabular}{|c|c|c|c|c|c|c|c|}
\hline
\textbf{Method} & \textbf{Type} & \textbf{Sup.} & \textbf{AP} & \textbf{AP50} & \textbf{AP75} & \textbf{AR1} & \textbf{AR10} \\
\hline
IDOL-BoxInst & Video & Box & 32.2 & 55.7 & 31.9 & 15.8 & 38.5 \\
PM-VIS & Video & Box & 37.5 & 62.6 & 37.5 & 16.8 & 43.9 \\
PM-VIS\textsuperscript{+} & Image & Pixel/Box & 19.3 & 39.3 & 17.4 & 11.9 & 28.7 \\
PM-VIS\textsuperscript{+} & Video & - & \textbf{22.1} & \textbf{43.2} & \textbf{20.2} & \textbf{12.7} & \textbf{30.4} \\
\hline
\end{tabular}
\label{tab:swinlOVIS}
\end{table}

As shown in Table~\ref{tab:swinlytvis2019} ,Table~\ref{tab:swinlytvis2021} and Table~\ref{tab:swinlOVIS} , we also conducted experiments using the Swin-L backbone network on the YTVIS2019, YTVIS2021, and OVIS datasets. It is evident that compared to using a weaker backbone network (ResNet50), the proposed model achieves a significant improvement in recognition capability.
When training the model using video data containing pseudo-masks, the model's recognition capability improves by 2.7\%, 2.9\%, and 2.8\% relative to using only image datasets, reaching 57.2\%, 52.9\%, and 22.1\%, respectively. Therefore, based on the experimental results, it is evident that the PM-VIS\textsuperscript{+} method achieves high recognition levels on three different datasets using backbone networks with different feature extraction capabilities (ResNet-50, Swin-L). This to some extent demonstrates the effectiveness of the proposed method.

\section{Conclusion}
To address the issue of VIS algorithms overly relying on costly annotated video data, this paper proposes a method using dynamic mask loss. It utilizes both image data with only bounding boxes and image data with instance-level pixel annotations to train the VIS model PM-VIS\textsuperscript{+}. Furthermore, to fully utilize existing video data resources, PM-VIS\textsuperscript{+}(Image) is used to infer and optimize unannotated video data, obtaining pseudo-labels with a certain level of credibility. Subsequently, these pseudo-labeled data are used to train the PM-VIS\textsuperscript{+} model, resulting in the VIS model PM-VIS\textsuperscript{+}(Video) with high recognition capability. It is evident that the proposed method achieves high VIS recognition rates without using any manually annotated video data. This approach provides a new perspective for the application research of VIS and reduces the cost of applying VIS methods.


\bibliographystyle{IEEEtran}  
\bibliography{egbib.bib}

\end{document}